\documentclass[9pt, conference]{IEEEtran}
\IEEEoverridecommandlockouts
\usepackage{cite}
\usepackage{amsmath,amssymb,amsfonts}
\usepackage{algorithmic}
\usepackage{graphicx}
\usepackage{textcomp}

\usepackage{booktabs}

\usepackage[table]{xcolor}
\usepackage{tikz}

\newcommand\copyrighttext{%
  \footnotesize \textcopyright 2025 IEEE.  Personal use of this material is permitted.  Permission from IEEE must be obtained for all other uses, in any current or future media, including reprinting/republishing this material for advertising or promotional purposes, creating new collective works, for resale or redistribution to servers or lists, or reuse of any copyrighted component of this work in other works.}
\newcommand\copyrightnotice{%
\begin{tikzpicture}[remember picture,overlay]
\node[anchor=south,yshift=10pt] at (current page.south) {\fbox{\parbox{\dimexpr\textwidth-\fboxsep-\fboxrule\relax}{\copyrighttext}}};
\end{tikzpicture}%
}

\def\BibTeX{{\rm B\kern-.05em{\sc i\kern-.025em b}\kern-.08em
    T\kern-.1667em\lower.7ex\hbox{E}\kern-.125emX}}
\begin{document}

\title{Rethinking Encoder-Decoder Flow Through Shared Structures\\
}

\author{\IEEEauthorblockN{1\textsuperscript{st} Frederik Laboyrie}
\IEEEauthorblockA{\textit{Samsung R\&D Institute UK (SRUK)} \\
London, England \\
f.laboyrie@samsung.com}
\and
\IEEEauthorblockN{2\textsuperscript{nd} Mehmet Kerim Yucel}
\IEEEauthorblockA{\textit{Samsung R\&D Institute UK (SRUK)} \\
London, England \\
mehmet.yucel@samsung.com}
\and
\IEEEauthorblockN{3\textsuperscript{rd} Albert Sa\`a-Garriga}
\IEEEauthorblockA{\textit{Samsung R\&D Institute UK (SRUK)} \\
London, England \\
a.garriga@samsung.com}

}

\maketitle
\copyrightnotice

\begin{abstract}
 Dense prediction tasks have enjoyed a growing complexity of encoder architectures, decoders, however, have remained largely the same. They rely on individual blocks decoding intermediate feature maps sequentially. We introduce banks, shared structures that are used by each decoding block to provide additional context in the decoding process. These structures, through applying them via resampling and feature fusion, improve performance on depth estimation for state-of-the-art transformer-based architectures on natural and synthetic images whilst training on large-scale datasets.
\end{abstract}

\section{Introduction}
\label{sec:intro}

State-of-the-art encoder-decoder structures for depth estimation utilise intermediate feature maps from pretrained transformer-based encoders \cite{dinov2} and decode them with comparatively simpler convolutional structures \cite{refinenet} \cite{yucel2021real}. These methods have achieved astonishing results lately \cite{depthanything} \cite{marigold} leveraging data labelling schemes, synthetic datasets and encoder pretraining regimes \cite{dinov2}. However, from an architectural standpoint, the state-of-the-art is essentially unchanged since the introduction of Dense Prediction Transformers \cite{dpt}. This in part due to the strong initialisation of Vision Transformers \cite{vit} (ViTs henceforth) with DINOv2 weights \cite{dinov2} being tied to this architecture, making it hard to deviate from this start point. Even so, multi-modal models with 1 billion+ parameters \cite{eva1} still utilise the same decoding process as introduced in \cite{dpt}.

This work aims to introduce an alteration to the encoder-decoder flow with the generation of banks. Rather than decoders purely operating on individual intermediate feature maps, here they operate additionally on shared tensors which contain information about all intermediate feature maps. In this way, earlier decoding blocks have more forward context, and likewise later decoding blocks have more backward context. The banks interact with the features directly as well as through a dual-interaction guided sampling procedure we introduce. Using our block design, we evaluate them on depth estimation, on which they are capable of gaining accuracy proportionally larger than the negligible increase in GFLOPS and parameters they introduce. Note that as a concept, they are not tied to either this task or a specific decoder and could well be applied to any dense prediction task which utilises multiple outputs from an encoder.

\begin{figure}[!t]
    \centering
    \includegraphics[width=\linewidth]{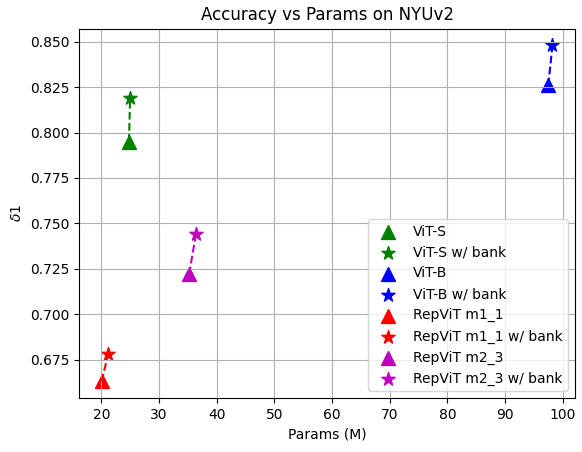}
    \vspace{-5mm}
    \caption{Accuracy ($\delta$1) vs parameters for architectures trained on large-scale dataset and tested on NYUv2. The introduction of our proposed banks can make a ViT-S based model almost match the performance of a ViT-B based model, whilst introducing a negligible amount of parameters. }
    \label{fig:accs}
    \vspace{-5mm}
\end{figure}

\section{Related Work}
\label{sec:relwork}

\begin{figure*}[t]
    \centering
    \includegraphics[width=\textwidth]{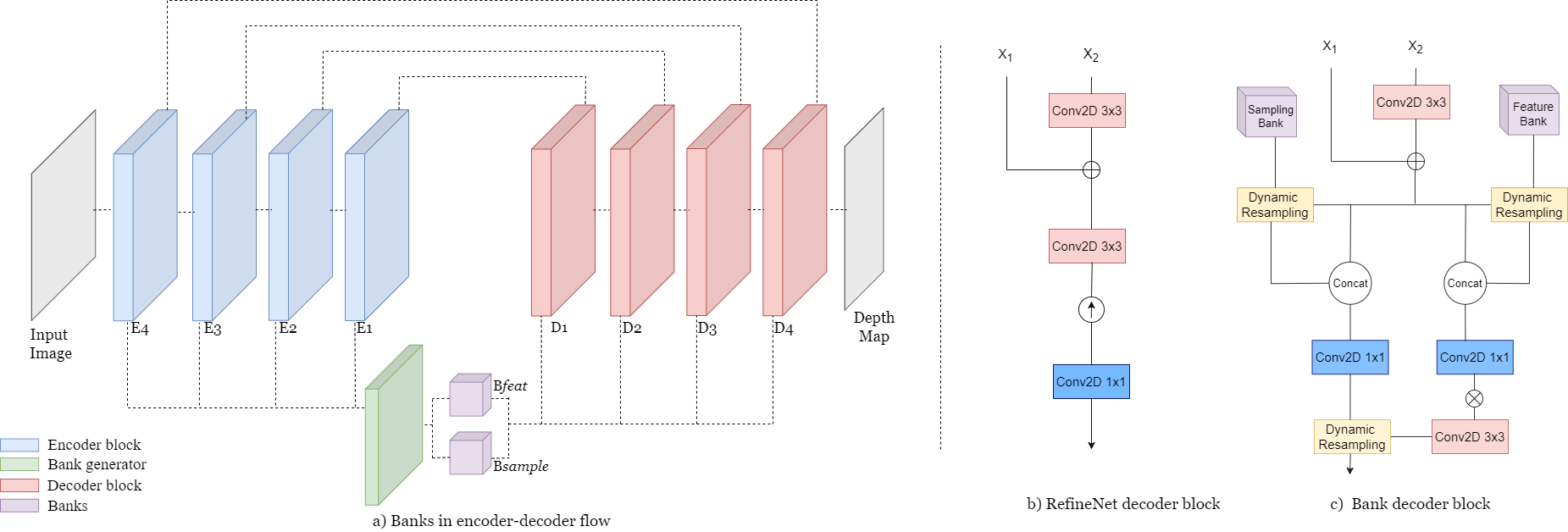}
    \caption{Encoder-decoder flow with banks (a) shown with RefineNet decoder block (b) and our block with banks (c). $\uparrow$ is bilinear upsampling. }
    \label{fig:unet}
        \vspace{-2mm}
\end{figure*}

This work focuses on monocular depth estimation, for which most recent work has been around unifying datasets \cite{midas} or utilizing unlabelled data \cite{depthanything} to be able to supervise this task well. Recent work is either tied to plain ViTs \cite{depthanything} or supports a family of transformers \cite{midas} \cite{zoedepth}. From an architectural standpoint, other work exists which focuses on diffusion models \cite{marigold} \cite{depthfm}. Additionally, research around bins-based methods focusing on metric depth \cite{binsformer} \cite{adabins} is prevalent, although for this work we focus on relative depth.

Research around encoders, specifically ViT-based ones, has either centered around introducing convolutional priors or altering the attention mechanism \cite{shvit} \cite{hilo}. Convolutional priors were originally seen in SWIN-based models \cite{swin} \cite{swinunet} via a sliding window mechanism, but are also seen in hybrid methods which introduce convolutional operations in the initial patchify operation in some manner consistently throughout the ViT encoder \cite{efficientformer, swiftformer, fastvit}. Finally, there are pure convolutional models which are at least transformer-informed \cite{repvit}.

Depth estimation decoders, along with decoders used in other dense prediction tasks, share the fact they are simple CNN-based decoders. In semantic segmentation, although differing to RefineNet, state-of-the art methods such as \cite{segformer} \cite{segmenter} utilise cheap pointwise convolutions and multi-layer perceptrons. In medical tasks, although pure transformer-based methods exist \cite{swinunet}, they still follow the same macro structure as defined in Section \ref{sec:method}.

As ViTs are the base for image-based tasks and have achieved remarkable performance on a variety of tasks, little research has been done to deviate largely from the paradigm. However, ViTs are not faultless and issues such as noisy feature maps are shared with all architectures which utilise positional encodings. Generic add-ons to tackle this issue were shown in \cite{denoisingvits}. This is where our work lies and aims to be a generic add-on for state-of-the-art ViT architectures used for dense prediction tasks.

\section{Method}
\label{sec:method}

\subsection{Formulation}
\label{sec:banks}

An image encoder {$\mathbf{E}=\langle E_1, \cdots, E_n\rangle$} is comprised of $n$ encoder blocks. It takes in an image $\textbf{I} \in \mathcal{R}^{H \times W \times 3}$ and each encoder block provides its respective output $\langle \mathcal{O}_{E_i}, \cdots, \mathcal{O}_{E_n}\rangle$. ViT \cite{vit}, as an encoder, outputs 4 intermediate features maps as standard. An image decoder {$\mathbf{D}=\langle D_1, \cdots, D_n\rangle$} is then comprised of $n$ decoding blocks and which provide their respective outputs $\langle \mathcal{O}_{D_i}, \cdots, \mathcal{O}_{D_n}\rangle$ and provide the final output depth map $\mathcal{O}_d{}_e{}_p{}_t{}_h{} \in \mathcal{R}^{H \times W}$. As shown in Figure \ref{fig:unet}, without the bank branch, if  $i$ denotes a given shared level between $\mathbf{D}$ and $\mathbf{E}$, each decoding block $\mathbf{D_i}$ will operate exclusively on feature map $\mathcal{O}_{E_i}$ and the output of the previous decoding block $\mathbf{D_i{}_-{}_1}$, excluding the very first decoding block. So we have:

\begin{align}
\mathcal{O}_{D_i} = \begin{cases}\mathbf{D_i} (\mathcal{O}_{E_i}, \mathcal{O}_{D_i{}_-{}_1}) & \text{if i $>$ 1,} \\
\mathbf{D_i}(\mathcal{O}_{E_i}) & \text{otherwise} \end{cases}
\end{align}

We reformulate this to accommodate our bank structures which are globally shared by all decoding blocks. The feature maps all go through the bank generator $\textbf{Z}$:

\begin{equation}
\langle B_s{}_a{}_m{}_p{}_l{}_e,B_f{}_e{}_a{}_t\rangle = \mathbf{Z} (\mathbf{E})
\end{equation}

where $\textbf{Z}$ is a pair of residual pointwise convolutional blocks generating $B_s{}_a{}_m{}_p{}_l{}_e$ and $B_f{}_e{}_a{}_t$ separately, which denote the sampling bank and feature bank, respectively. Their application is shown in Figure \ref{fig:unet}C and described in more detail in Sections \ref{sec:featbanks} and \ref{sec:samplbanks}. With these structures the decoding process is reformulated as:

\begin{align}
\mathcal{O}_{D_i} = \begin{cases}\mathbf{D_i} (\mathcal{O}_{E_i}, \mathcal{O}_{D_i{}_-{}_1}, B_s{}_a{}_m{}_p{}_l{}_e, B_f{}_e{}_a{}_t) & \text{if i $>$ 1,} \\
\mathbf{D_i}(\mathcal{O}_{E_i}, B_s{}_a{}_m{}_p{}_l{}_e, B_f{}_e{}_a{}_t) & \text{otherwise} \end{cases}
\end{align}

A given decoding block $\mathbf{D_i}$ now no longer purely operates singularly on adjacent feature maps relative to their position in the decoding process, but also take as input the shared bank objects.

\begin{figure}[t]
    \centering
    \includegraphics[width=\linewidth]{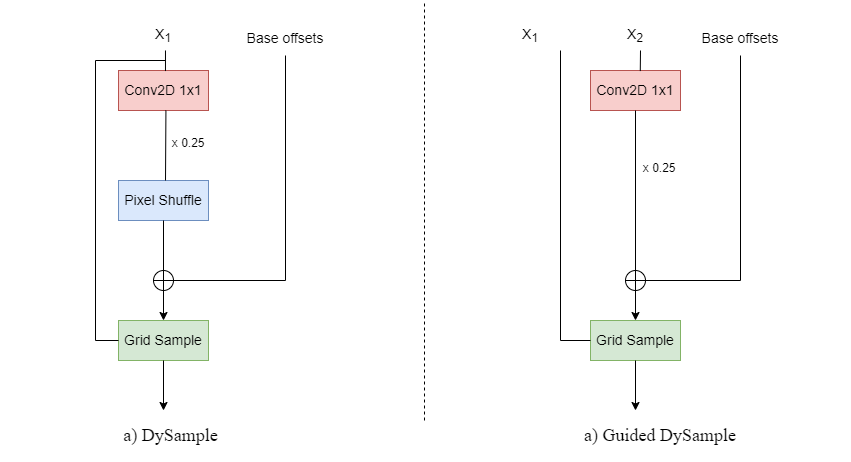}
    \caption{DySample (a) reconfigured to support guided sampling via reference tensors (b).}
    \label{fig:dys}
    \vspace{-4mm}
\end{figure}

\subsection{Banks}
\subsubsection{Feature Banks}
\label{sec:featbanks}

If a bank $B$ which has interacted with a feature map $\mathbf{X}$ is denoted as $B^\prime$,
 we can formulate this via a concatenation and joint convolution to form resultant feature map $X^\prime$ as:

\begin{equation}
B^\prime = conv(concat(B, X))
\end{equation}
\begin{equation}
X^\prime = X \times B^\prime
\end{equation}

In this way, the bank reweights the features via element-wise multiplication. Note that in our case $\mathbf{X}$ would be the resultant fusion of $\mathbf{X_1}$ and $\mathbf{X_2}$, as shown in Figure \ref{fig:unet}C.

\begin{figure*}[!t]
    \centering
    \includegraphics[width=\textwidth]{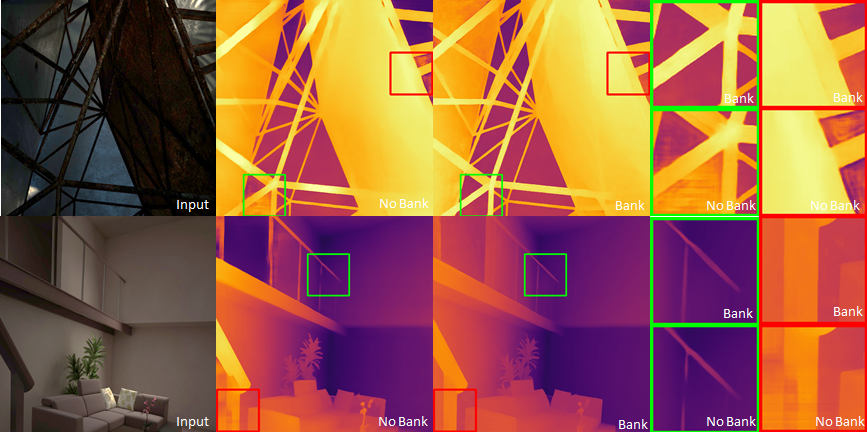}
    \caption{Qualitative results of RepViT m2\_3 on HyperSim dataset. The usage of banks recovers artifacts and improves geometric accuracy of edges. }
    \label{fig:quali}
    \vspace{-3mm}
\end{figure*}

\subsubsection{Sampling Banks}
\label{sec:samplbanks}

The other mechanism through which banks interact with the feature maps is the dynamic upsampling of the fused feature maps entering the decoding block. In RefineNet \cite{refinenet}, the output feature map undergoes bilinear upsampling. In our blocks, the sampling bank $B_s{}_a{}_m{}_p{}_l{}_e$  is used as a guidance tensor for a dynamic upsampling based on DySample \cite{dysample}. In DySample \cite{dysample}, feature maps are convolved over, which is followed by an upsampling pixel shuffle operation to generate residuals. These residuals are added onto base offsets, which are initialized to mimic bilinear upsampling. 

In our blocks we introduce a guided version, in which a guidance tensor of the target resolution is convolved over to directly generate the offset residuals. In this way, we readily support non-integer sampling factors and downsampling. With non-integer sampling factors, we can naturally support odd-dimensional feature maps, which occurs in certain ViT configurations \cite{vit}. Downsampling also allows a further interaction of $B_s{}_a{}_m{}_p{}_l{}_e$ and feature maps within decoding blocks, that is the downsampling of the bank to the target resolution of the decoding block. 

Downsampling is necessary as banks are generated at the largest resolution of the intermediate feature maps. Smaller intermediate feature maps are bilinearly upsampled to that larger resolutions before being fused. In order to utilize sampling banks in earlier decoding blocks. it is necessary to downsample the banks in order for them to guide the upsampling of significantly smaller feature maps. This is because the interpolation factor is typically 2. Instead of downsampling them via a non-dynamic interpolation algorithm, in our block we downsample the blocks with the feature maps a guidance tensor.

In a general formulation, we can define a guided sampling function $\mathbf{GS}\uparrow/\downarrow$ as:

\begin{equation}
GS\uparrow/\downarrow(X_i{}_n, X_r{}_e{}_f): \mathbb{R} \rightarrow \mathbb{R}
\end{equation}

in which $X_i{}_n$ and $X_r{}_e{}_f$ denote the input and reference tensors, respectively. $\uparrow$ and $\downarrow$ denote whether the output tensor is upsampled or downsampled, for clarity. Then, in a general formulation with a given feature map $\mathbf{X}$, a given bank $B_s{}_a{}_m{}_p{}_l{}_e$ and their respective output $\mathcal{O}$ we have:

\begin{equation}
O = GS\uparrow(X, GS\downarrow(B, X))
\end{equation}

In our block, the feature maps used in this formulation are either the output from the feature bank interaction described in Section \ref{sec:samplbanks}, or again the initial fusion of $X_1$ and $X_2$ as shown in Figure \ref{fig:unet}C.

\section{Experiments}
\label{sec:experiments}

\begin{table*}[t]
    \caption{Results from training on large-scale mixed natural and synthetic dataset.}

\footnotesize
    \centering
    \setlength\tabcolsep{1.75mm}
    \renewcommand{\arraystretch}{1.05}
    \begin{tabular}{rccccccccccccc}
    \toprule
    ~ & ~ & \multicolumn{4}{c}{NYUv2\cite{nyu}} & \multicolumn{4}{c}{Mannequin Challenge\cite{mc}} & \multicolumn{4}{c}{IRS\cite{irs}} \\

    \cmidrule(lr){3-6} \cmidrule(lr){7-10} \cmidrule(lr){11-14}

~ & ~ & \multicolumn{2}{c}{No Bank} & \multicolumn{2}{c}{Bank} & \multicolumn{2}{c}{No Bank} & \multicolumn{2}{c}{Bank} & \multicolumn{2}{c}{No Bank} & \multicolumn{2}{c}{Bank} \\

\cmidrule(lr){3-4} \cmidrule(lr){5-6} \cmidrule(lr){7-8} \cmidrule(lr){9-10} \cmidrule(lr){11-12} \cmidrule(lr){13-14}

    {Encoder} & {Initial Weights} &  {$\delta_1$$\uparrow$} & {AbsRel$\downarrow$} & {$\delta_1$$\uparrow$} & {AbsRel$\downarrow$} &  {$\delta_1$$\uparrow$} & {AbsRel$\downarrow$} & {$\delta_1$$\uparrow$} & {AbsRel$\downarrow$}&  {$\delta_1$$\uparrow$} & {AbsRel$\downarrow$} & {$\delta_1$$\uparrow$} & {AbsRel$\downarrow$}  \\
    \midrule
    ViT-S  & DINOv2~\cite{dinov2} & 0.795 & 0.213 & \textbf{0.819} & \textbf{0.185} & 0.767 & 0.272 & \textbf{0.769} & \textbf{0.257} & 0.761 & 0.272 & \textbf{0.790} & \textbf{0.242}  \\

    ViT-B & DINOv2~\cite{dinov2} & 0.826 & \textbf{0.167} & \textbf{0.848} & 0.175 & 0.797 & \textbf{0.219} & \textbf{0.802} & 0.228 & 0.818 & 0.209 & \textbf{0.819} & \textbf{0.207} \\

    \midrule

    RepViT m1\textunderscore1 & Imagenet21k~\cite{imagenet21k} & 0.663 & 0.369 & \textbf{0.678} & \textbf{0.325} & 0.646 & 0.474 & \textbf{0.660} & \textbf{0.412} & 0.617 & 0.487 & \textbf{0.633} & \textbf{0.436} \\

    RepViT m2\textunderscore3 & Imagenet21k~\cite{imagenet21k} & 0.722 & 0.3075 & \textbf{0.744} & \textbf{0.253} & 0.678 & 0.416 & \textbf{0.688} & \textbf{0.368} & 0.687 & 0.368 & \textbf{0.702} & \textbf{0.348} \\

    \bottomrule
    \end{tabular}
    \vspace{2mm}
    \label{tab:fulltrain}
        \vspace{-3mm}
\end{table*}

\begin{table*}[t]
\caption{Results from training and evaluating on HyperSim.}
\footnotesize
    \centering
    \setlength\tabcolsep{1.75mm}
    \renewcommand{\arraystretch}{1.05}
    \begin{tabular}{rccccccccccccc}
    \toprule
    ~ & ~ & \multicolumn{6}{c}{No Bank} & \multicolumn{6}{c}{Bank} \\

    \cmidrule(lr){3-8} \cmidrule(lr){9-14}

    {Encoder} & {Initial Weights} &  {$\delta_1$$\uparrow$} & $\delta_2$$\uparrow$ & $\delta_3$$\uparrow$ & {AbsRel$\downarrow$} & {RMSE$\downarrow$} & log10$\downarrow$ &  {$\delta_1$$\uparrow$} & $\delta_2$$\uparrow$ & $\delta_3$$\uparrow$ & {AbsRel$\downarrow$} & {RMSE$\downarrow$} & log10$\downarrow$ \\
    \midrule
    ViT-S  & DINOv2~\cite{dinov2} & \textbf{0.918} & 0.962 & \textbf{0.978} & 0.147 & 0.055 & 0.039 & \textbf{0.918} & \textbf{0.963} & \textbf{0.978} & \textbf{0.145} & \textbf{0.054} & \textbf{0.038} \\

    ViT-B & DINOv2~\cite{dinov2} & 0.923 & 0.964 & \textbf{0.979} & 0.134 & 0.528 & 0.039 & \textbf{0.924} & \textbf{0.968} & 0.978 & \textbf{0.138} & \textbf{0.524} & \textbf{0.038} \\

    \midrule

    RepViT m1\textunderscore1 & Imagenet21k~\cite{imagenet21k} & 0.897 & 0.953 & 0.974 & 0.173 & 0.063 & 0.047 & \textbf{0.902} & \textbf{0.958} & \textbf{0.978} & \textbf{0.161} & \textbf{0.061} & \textbf{0.044} \\
    RepViT m2\textunderscore3 & Imagenet21k~\cite{imagenet21k} & 0.919 & 0.965 & 0.980 & 0.134 & 0.052 & 0.039 & \textbf{0.926} & \textbf{0.967} & \textbf{0.981} & \textbf{0.129} & \textbf{0.049} & \textbf{0.035} \\

    \bottomrule
    \end{tabular}

    \label{tab:hypersimtrain}
            \vspace{-3mm}
\end{table*}

\newcommand{\gr}{\rowcolor[gray]{.95}}

\begin{table*}[htp]
\caption{Footprint analysis of banks on our model suite.}
\centering
\small
\addtolength{\tabcolsep}{1.5pt}
\begin{tabular}{lccccc}
\toprule
~ & ~ & \multicolumn{2}{c}{No Bank} & \multicolumn{2}{c}{Bank} \\
\cmidrule(lr){3-4} \cmidrule(lr){5-6}
Models & Bank Channels  & GFLOPs & Parameters(M) & GFLOPs  & Parameters(M)  \\
\midrule 
ViT-S & 64 & 80.30  & 24.79 & 83.79  & 24.99 \\
ViT-B & 128 & 311.0  & 97.47 & 331.7  & 98.20\\
\midrule
RepViT m1\textunderscore1  & 256 & 253.2  & 20.10 & 261.4  & 21.11\\
RepViT m2\textunderscore3  & 256 & 289.6  & 35.21 & 299.7  & 36.42\\

\bottomrule
\end{tabular}

\label{tab:footprint}
            \vspace{-3mm}
\end{table*}

\begin{table*}[htp]
\caption{Ablation results for bank design.}
\centering
\small
\addtolength{\tabcolsep}{1.5pt}
\begin{tabular}{lccc}
\toprule
Model + bank types & NYUv2 \cite{nyu} $\delta_1$  & GFLOPs  & Parameters(M)\\
\midrule
\rowcolor[gray]{.95} 
ViT-S & 0.795 & 80.30  & 24.79 \\
+ feature & 0.808 & 83.74  & 24.94\\
+ feature + upsampling  & 0.799 & 83.74  & 24.99\\
+ feature + upsampling + downsampling  & 0.819 & 83.79 & 24.99\\

\bottomrule
\end{tabular}
\vspace{2mm}
\label{tab:ablate}
            \vspace{-3mm}
\end{table*}

\subsection{Training procedure}
\label{sec:train prod}
Our experimentation procedure involves taking ViT \cite{vit} and RepVit \cite{repvit} encoders pretrained on either Imagenet-21k \cite{imagenet21k} or DINOv2 weights \cite{dinov2} as start points. We then train them with freshly initalized decoders with a starting learning rate of 5e-5 using L1 loss exclusively. Further experimentation with depth specific losses to optimize performance was beyond the scope of this work, which is to prove the efficacy of bank structures at a general architectural level. 

We gather two main results, the first being training on HyperSim dataset for 500k iterations. As they are rendered images, not natural images, we use the synthetic labels provided by the renderer as ground truth. Second, we train on a mix of large open datasets includes Open Images\cite{openimages}, Mannequin Challenge\cite{mc}, NYUv2\cite{nyu} and IRS\cite{irs} which together are comprised of over 1 million natural and synthetic images. We train and evaluate on DepthAnythingV2 pseudo-labels as opposed to the ground truth in the evaluation datasets. This provides more consistency between the training data and evaluation labels which is beneficial for evaluating our idea. We evaluate using standard depth metrics $\delta_1$, $\delta_2$, $\delta_3$, AbsRel, RMSE, and log10 \cite{dpt} \cite{depthanything}. These are all pixel-level error metrics and are thresholded or reweighted metrics heavily related to mean error.

\subsection{Results}
\label{sec:resultsv2}

Results in Tables \ref{tab:fulltrain} and \ref{tab:hypersimtrain} show that the introduction of banks are able to cause increase in all metrics across the board, albeit with a few exceptions. ViT-B, despite gains in $\delta$1 shows modest deterioration in the AbsRel metric on Mannequin Challenge and IRS. There is also a minor deterioration in $\delta3$ for ViT-B on HyperSim. We highlight that ViT-S is able to bridge a large part of the gap with ViT-B on NYUv2 with respect to $\delta$1 despite still having significantly fewer parameters, as shown in Figure \ref{fig:accs}. 

When training on smaller-scale datasets in the case of Hypersim \cite{hypersim}, in Table \ref{tab:hypersimtrain} the results on the fully convolutional RepViT are more consistent and slightly more pronounced. This is expected as transformer-based architectures generally require a more significant amount of data to gain their full potential. 

In Table \ref{tab:footprint}, we show the effect of introducing banks to decoders on GFLOPs and parameter count. This provides context for the accuracy increases shown. The introduction of banks in terms of parameters is negligible for ViTs at less than 1\% increase and around 3-5\% for RepViTs. The increase in GFLOPs is around 3\% for RepViT and 4-6\% for ViT architectures. This increase in GFLOPs is significantly less than increase the class denomination of the encoder. ViT-B represents around a 300\% increase in parameters and GFLOPs from ViT-S. It is clear that the gains in accuracy are cheaper via introducing banks, as opposed to scaling the encoder via a class denomination.

Despite banks failing to show strong metric increases in every case, specifically HyperSim, the qualitative improvements are consistent. we show in Figure \ref{fig:quali} the ability for banks to recover artifacts even in evaluations which are modest in terms of metric improvements. The introduction of banks can remove block artifacts, semantic error as well as geometric inaccuracies purely from an architectural level.

\subsection{Ablations}
\label{sec:ablations}
As ViTs are our target architecture, we ablate on them to settle on block design and understand the effect of each component.
Ablations were performed using DepthAnythingV2 \cite{depthanything} pseudo-labels with the same training procedure as in Section \ref{sec:train prod} in Table \ref{tab:fulltrain}. We remove the output convolution from our block design in place of our guided sampling operation. For this reason, blocks with only feature banks have fewer GLOPS than the baseline. 

The metric results are shown in Table \ref{tab:ablate} and show the effect of using feature banks exclusively, as well as two variations of the application of the sampling banks. Feature banks on their own are capable of providing gains in performance, even with the removal of the output convolution. Upsampling banks are a detriment to metric performance unless they are dynamically downsampled with guidance by the feature maps.

We wish to emphasise here that the introduction of sampling banks even without the dynamic downsampling brought about the majority of qualitative gains, despite a metric deterioration. We observed sharper edges and absence of ghosting artifacts apparent in models which did not utilise the sampling banks. This highlights the power of the guided sampling procedure when applied in either manner and at a cheaper cost, when compared to the application of feature banks.

\section{Conclusion}
\label{sec:conclusion}
We show that shared banks that are used in conjunction by each decoding block improve transformer-based models performance in depth estimation, quantitatively and qualitatively. We demonstrate the guided sampling in particular, via bank interactions, can cause most significant gains.

To follow up on this work, there exists clear paths in developing the block design and bank generation further. The interactions and generation could be much heavier to to understand the maximum potential. In addition, the guided sampling developed to make use of the banks could be developed in its own right. Perhaps most importantly, there exists avenues for applying banks to other dense prediction tasks which utilises intermediate features maps, such as semantic segmentation.

\end{document}